# Panning for gold: Lessons learned from the platform-agnostic automated detection of political content in textual data

Mykola Makhortykh*, Ernesto de León*, Aleksandra Urman**, Clara Christner***, Maryna Sydorova*, Silke Adam*, Michaela Maier***, and Teresa Gil-Lopez****

* Institute of Communication and Media Studies, University of Bern

** Social Computing Group, University of Zurich

*** Institute for Communication Psychology and Media Education, University of Koblenz-Landau

**** Department of Communication, University Carlos III of Madrid

## Author Note

This is the preprint version of the "Panning for gold: Lessons learned from the platform-agnostic automated detection of political content in textual data" article. The article is written within the project “Reciprocal relations between populist radical-right attitudes and political information behaviour: A longitudinal study of attitude development in high-choice information environments” led by Silke Adam (University of Bern) and Michaela Maier (University of Koblenz-Landau) and sponsored by the SNF (100001CL_182630/1) and DFG (MA 2244/9-1).

**Authorship statement**

*Funding acquisition*: Silke Adam, Michaela Maier

*Project administration*: Silke Adam, Michaela Maier

*Research design for this paper:* Mykola Makhortykh, Ernesto de León, Silke Adam, Michaela Maier, Aleksandra Urman, Teresa Gil-Lopez, Clara Christner

*Development of dictionaries*: Mykola Makhortykh, Ernesto de León, Silke Adam, Michaela Maier, Aleksandra Urman, Maryna Sydorova

*Development of SML models*: Ernesto de León

*Development of NN models*: Mykola Makhortykh

*Training data acquisition*: Ernesto de León, Mykola Makhortykh

*Validation dataset preparation*: Clara Christner, Silke Adam, Michaela Maier

*Server-side implementation for model deployment / training*: Mykola Makhortykh, Maryna Sydorova

*Model validation*: Mykola Makhortykh, Ernesto de León, Maryna Sydorova, Aleksandra Urman

*Paper writing*: Mykola Makhortykh, Ernesto de León

*Editing*: Silke Adam, Michaela Maier

# Panning for gold: Lessons learned from the platform-agnostic automated detection of political content in textual data

**Abstract**: The growing availability of data about online information behaviour enables new possibilities for political communication research. However, the volume and variety of these data makes them difficult to analyse and prompts the need for developing automated content approaches relying on a broad range of natural language processing techniques (e.g. machine learning- or neural network-based ones). In this paper, we discuss how these techniques can be used to detect political content across different platforms. Using three validation datasets, which include a variety of political and non-political textual documents from online platforms, we systematically compare the performance of three groups of detection techniques relying on dictionaries, supervised machine learning, or neural networks. We also examine the impact of different modes of data preprocessing (e.g. stemming and stopword removal) on the low-cost implementations of these techniques using a large set (n = 66) of detection models. Our results show the limited impact of preprocessing on model performance, with the best results for less noisy data being achieved by neural network- and machine-learning-based models, in contrast to the more robust performance of dictionary-based models on noisy data.

**Introduction**

The emergence of the present high-choice media environment (van Aelst et al. 2017) has been accompanied by an unprecedented expansion of the volume of politics-related content and the formats in which it is consumed (e.g. microblogs and online newspapers; Mukerjee et al. 2018). To understand political information consumption in such an environment, we need new approaches to trace individual interactions with content online (e.g. clickstream or web-tracking data; Christner et al. 2021; Makhortykh et al. 2021), but also novel techniques to analyse the content (e.g. to detect the presence of politics-related information; Wojcieszak et al. 2021).

To date, most studies have relied on the information source (e.g. web domain type) to distinguish the information individuals engage with online (e.g. Dvir-Gvirsman et al. 2016; Stier et al. 2020). While some studies have also looked at actual content, the usual focus is on a single type of data, such as journalistic articles (de León and Trilling 2021), tweets (de Mello Araujo and Ebbelaar 2018), and Anglophone texts. While such a specific focus is common in natural language processing (NLP) research, to which the "no free lunch" theorem[1] is applicable, it limits the possibilities for studying engagements with (political) content that occur across multiple platforms.

In this article, we discuss the lessons learned from our work on the automated detection of politics-related information in multi-platform German textual content. Specifically, we compare the performance of three groups of detection techniques based on dictionaries, supervised machine learning (SML), and neural networks (NN) across three validation datasets with varying degrees of text "noise" (Agarwal et al. 2007). In doing so, we also discuss how the performance of the low-cost implementations of these techniques is affected by different modes of text preprocessing. We argue that despite the multiple challenges associated with developing platform-agnostic content analysis approaches, pursuing this task is important for realising the opportunities enabled by new forms of cross-platform passive data collection (Stier et al. 2020a) and advancing research on diverse communication phenomena, ranging from selective exposure (Guess et al. 2021) to media effects (de León et al. 2022) to news consumption inequalities (Merten et al. 2022).

[1] The "no free lunch" theory refers to the idea that more universal computational approaches will always underperform compared with more specific/narrow ones (Ho and Pepyne 2002).

The rest of the paper is organised as follows. First, we briefly review the existing research on the automated detection of political content and the impact of text preprocessing on automated content analysis techniques. We then introduce the methodology used to compare the performance of different detection models and different modes of preprocessing. After this, we share our findings about the models' performances across the three validation datasets and discuss the implications of the lessons learned through this process, together with the study's limitations and directions for future research.

## Literature review

### *Automated detection of politics-related information*

While the ability to detect politics-related information is important for studying communication in online environments, the practical realisation of this task is complicated by the different formats in which such information can appear. Existing studies often address this challenge by assuming that all content coming from a specific source (e.g. a news website; Dvir-Gvirsman et al. 2016) or of a specific type (e.g. political manifestos; Benoit et al. 2016) is related to politics. However, the rise of multi-purpose platforms (e.g. social media; de Mello Araujo and Ebbelaar 2018), where individuals can engage with both politics- and non-politics-related information, and the new formats through which politics-related information can be disseminated prompt the need for approaches looking at the presence of such information on the content level. These approaches rely on one of three groups of NLP techniques: dictionary, SML, and neural network-based approaches.

*Dictionary-based techniques* utilise lists of terms related to the specific construct to be detected (Dun et al. 2021). Compared with more complex SML- and NN-based techniques, the main principle of dictionary-based approaches is rather simple: if the piece of content contains a certain number of terms present in the dictionary, then it is classified as political. The simplicity and transparency of dictionaries have contributed to their active use for detecting political content, based on the presence of certain features like politicians' or parties' names (Barberá et al. 2021; Sang and Bos 2012) or terms associated with a specific politics-related phenomenon (e.g. migration; Heiss and Matthes 2020). Such dictionaries have also been used to detect political information within content coming from different platforms, such as news websites (Boumans and Trilling 2016) and social media (Sang and Bos 2012).

*SML-based techniques* rely on the likelihood of individual terms being representative of a specific content category (Boumans and Trilling 2016). Using manually annotated corpora (e.g. of sentences or documents), these techniques employ statistical models (Hamoud et al. 2018) to predict whether a specific piece of content has certain features (e.g. those related to politics). Despite being less transparent than dictionaries, SML-based techniques involve less preparatory work: instead of a list of terms representative of a particular issue, they require just a set of labelled data (e.g. based on manual annotation or metadata; de León et al. 2021; Stier et al. 2021). Together with their higher performance compared with dictionaries[2], their ease of use has contributed to the growing application of SML for politics-related information detection (e.g. on social media; de Mello Araujo and Ebbelaar, 2018).

Compared with SML, *NN-based techniques* make better use of contextual information and are more capable of processing unstructured data (Chang and Masterson 2020). In particular, convolutional neural networks (CNN) and long short-term memory networks (LSTM) have shown promising performance in text classification (Luan and Lin 2019). The effectiveness of NNs in recognising sequential patterns is further amplified by transformer networks, such as bidirectional encoder representations from transformers (BERT), which rely on a substantial volume of contextual information (Devlin et al. 2019). Although notorious for their lack of transparency (Kim et al. 2020), recent studies have demonstrated the substantial advantages of NNs in terms of performance on political detection tasks for social media (Rao et al. 2016) and journalistic content (Kulkarni et al. 2018).

Despite relying on different techniques to detect politics-related information, the majority of studies noted above share a common feature: the tendency to focus on content coming from a single platform (e.g. Twitter; Rao et al. 2016). Such a monoplatform focus limits the applicability of existing approaches to large datasets dealing with cross-platform engagement with online content. The growing availability of such datasets which are provided, for instance, by web-tracking studies[3], stresses the importance of more platform-agnostic approaches for detecting politics-related information.

*Preprocessing and automated content analysis*

[2] For the systematic comparison of the performance of these different techniques in another common content analysis task (i.e. sentiment detection), see van Atteveldt et al. (2021).

[3] An example of the volume of data used by these studies is given by Stier et al. (2021), who worked with 150 millions visits from thousands of different web domains. Another study by Wojcieszak et al. (2021) relied on 36.8 millions visits, with a median number of different domains visited per user of more than 800.

Preprocessing decreases the amount of noise in textual data by reducing the complexity of textual features (Grimmer and Stewart 2013), which is particularly important for dealing with multi-platform data, where the amount of noise is higher compared with monoplatform data. Some basic modes of preprocessing include letter lowercasing, punctuation removal, and the exclusion of repeating characters (HaCohen-Kerner et al. 2020). More complex modes deal with stopword removal (i.e. the removal of very common words or those with little meaning, such as “the”); stemming, where words are stripped to their base by removing verb and adverb suffixes (e.g. “ed” and “ly”); and lemmatisation, where inflicted versions of words are converted to their neutral lemma (e.g. “am” and “is” become “be”).

Despite the increasing number of studies conducting systematic analyses of the effects of preprocessing on the performance of automated content analysis approaches (e.g. Denny and Spirling 2018; HaCohen-Kerner et al. 2020), the choice of an optimal mode of preprocessing remains a challenging task. Its complexity can be attributed to several factors. First, the majority of studies compare the impact of preprocessing within a specific group of analytical techniques (e.g. SML; HaCohen-Kerner et al. 2020) and rarely examine the variation between different groups of techniques (e.g. SML and NN). Second, while the effects of preprocessing are influenced by the contextual factors associated with the task (e.g. the language of the dataset), most research focuses on Anglophone textual data, thus limiting the possibilities for investigating this impact. Third, there can be multiple implementations of the same preprocessing mode; while the impact of such differences can be marginal (e.g. different stemmers resulting in less than a 0.01 difference in accuracy scores; Bounabi et al. 2019), it nevertheless causes variation in the technique’s performance.

Under these circumstances, many studies argue for the use of less complex modes of preprocessing that require fewer computational resources but nevertheless improve model performance. For instance, HaCohen-Kerner et al. (2020) compared the impact of simpler forms of preprocessing on the performance of three SML models for Anglophone data and found that stopword removal resulted in the most consistent performance improvement for two out of three of the models. Similarly, the beneficial effect of stopword removal was observed in a study examining the effects of preprocessing on SML-based techniques for Czech data (Toman et al. 2006). However, in some cases, such as when dealing with content in Hebrew (HaCohen-Kerner et al. 2018) or corpora with few stopwords (e.g. spam mails; Méndez et al. 2005), the removal of stopwords actually worsened the model performance.

Compared with simple modes of preprocessing, such as stopword removal or the reduction of repeated characters, more complex modes (e.g. lemmatisation) enable more feature reduction and thus can provide larger performance increases for techniques affected by data noise (e.g. SML-based ones; Denny and Spirling 2018). In practice, however, the effect of complex modes of preprocessing turns to be rather ambiguous: in some use cases, particularly those dealing with SML, stemming (Gonçalves et al. 2010) or lemmatisation with stopword removal (El Kah and Zeroual 2021) result in performance improvements. In other cases, these modes of preprocessing result in marginal improvement (Song et al. 2005) or an actual drop in performance (Toman et al. 2006).

This ambiguous effect of more complex forms of preprocessing is particularly pronounced in the case of NN-based techniques, which sometimes benefit from higher noise (e.g. in the form of stopwords) that enables more possibilities to understand contextual relationships within the corpus. Maulana and Maharani (2021) showed that in the case of disaster-related Twitter content in English, stemming and stopword removal led to a decrease in the performance of the BERT because of the elimination of text features. At the same time,

Konstantinov et al. (2020) found that for Russian language classification tasks using BERT, lemmatisation enabled performance improvement, whereas stemming did not.

## Methodology

### *Political content detection*

We understand political content as material mentioning political actors/institutions and societally relevant issues that are tackled by political means (e.g. the economy; for more information on this definition, see Appendix A1). Because of our use case (see Appendix A2), we were particularly interested in Swiss/German actors, as well as the political issues relevant to these two countries. Although they have implications for the direct reuse of the detection models we developed, we expect that the observations generated through our model comparison will be applicable to a broad range of contexts.

To detect political content, we used three groups of techniques: SML, NN, and dictionary-based ones[4]. We were interested in the low-cost implementation of these techniques, in particular in terms of minimising the resources required to obtain data for model training and conducting the actual training (e.g. time- and processing-wise). Such an interest is attributed to our assumption that many academic projects might have limited financial/time resources for in-house NLP technique development; hence, we wanted to compare the performance of detection techniques under these unfavourable circumstances.

For SML, we used five models: logistic regression (LR), passive aggressive (PA), Bernoulli naive Bayes (BNB), multinomial naive Bayes (MNB), and stochastic gradient descent (SGD). These models differ in complexity, with some being based on simple Bayesian probabilistic modelling (e.g. BNB) and others relying on more advanced incremental learning principles (e.g. PA). All SML models were trained using the Scikit-Learn package for Python (Pedregosa et al. 2011).

---

[4] The trained SML models and dictionaries are available via OSF repository (https://osf.io/e8xtb/?view_only=0c58144e1769492cb32dd2d650062534). The trained NN models are available via the Harvard Dataverse repository (https://dataverse.harvard.edu/dataset.xhtml?persistentId=doi:10.7910/DVN/8Q5FPE).

For NNs, we used three models: CNN, LSTM, and BERT. CNNs have low computational costs compared with other types of NNs and focus on high-level features, while LSTMs place major emphasis on term sequences. Finally, BERT (Devlin et al. 2019) is a transformer model characterised by high computational costs but also more advanced capabilities for processing sequential data, along with an extensive awareness of contextual relationships between words attributed to it being pretrained on a large text corpus.

To train the CNN and LSTM models, we used the Python Tensorflow library (Abadi et al. 2016), whereas for BERT, we relied on a pretrained model for the German language from HuggingFace (2020). Because of our interest in low-cost detection approaches, we used simple CNN and LSTM architectures (see Appendix 3) with five learning epochs and 256 embedding dimensions. For BERT, we used three epochs because of the higher computational costs of model training and the need for no additional fine-tuning and tested a series of probabilities for the “political” label to be assigned. Based on the F1 scores achieved per probability (see Appendix 4), we opted for 0.15 probability, which resulted in the highest F1 score.

As training data for the SML- and NN-based models, we used a set of 4,023 articles crawled from German and Swiss journalistic media (i.e. *Blick*, *Bild*, and *Süddeutsche Zeitung*). To minimise the efforts required to manually annotate the data, we relied on metadata-based annotation in the form of journalistic tags (i.e. news categories such as “politics” or “sport”; Stier et al. 2021). These tags were used as labels to divide the crawled data into 1,523 political and 2,500 non-political articles, which were then used for model training based on an 80–20 train–test split. While the presence of such out-of-the-box annotation is a substantial advantage to using journalistic content as training data, it can also make the resulting detection models less effective for processing non-journalistic content.

For dictionaries*,* we used three models: Di-CAP, Di-LL, and Di-CAP-LL. The Di-CAP dictionary is made up of terms from the German codebook for the Comparative Agendas Project (CAP; Bevan 2019). The codebook is used to label political themes in Germany (e.g. economy or foreign politics) and includes key terms related to them. We also added a theoretically conceptualised list of terms for topics underrepresented in the CAP codebook (i.e. elections and ecology), together with a list of political actors' names in Germany/Switzerland (e.g. members of parliament) and G20/EU countries (e.g. presidents and vice-ministers).

The Di-LL dictionary is based on the same set of 4,023 journalistic articles that were used for the SML- and NN-based models. The two subsets of articles—political and non-political—were transformed into bags of words. Then, we used log-likelihood keyword analysis (Pojanapunya and Todd 2018) to identify terms that were overrepresented in the political subset. Following existing studies (e.g. de Schryver 2012), we created two sub-dictionaries consisting of the top 100 and top 1,000 terms (according to log-likelihood scores) and then compared their performance across three validation datasets using the no-preprocessing option (see below). Based on the average F1 scores across the validation datasets, the 100-term option demonstrated better performance and was thus used in the study.

Finally, the Di-CAP-LL dictionary combines terms from the Di-CAP and Di-LL dictionaries with filtered-out duplicates. The assumption here is that the combined dictionary might outperform its components by bringing together the theoretically informed set of terms (Di-CAP) and the empirically driven set of terms that were most common in politics-related journalistic articles (Di-LL).

A major difference in using dictionaries compared with SML- and NN-based techniques is that, by default, dictionaries do not provide a binary label (i.e. an indicator of whether the document is related to politics). Instead, dictionaries allow for the identification of the number of politics-related terms within a document that has then to be translated into a binary label. To address the possible variation in the length of documents coming from different platforms, we relied not on the absolute numbers, but on the ratio between the number of politics-related unique terms to the overall number of unique terms per document.

After calculating all the ratios per validation dataset, we then used each of these ratios as a possible threshold for assigning a political label to all the documents in the respective validation dataset and calculated the resulting average F1 scores. Then, we chose the three thresholds that achieved the maximum F1 score for the three validation datasets (i.e. the one threshold per dataset) and applied each of these thresholds to all three datasets to identify a single threshold that would demonstrate the most consistent performance. This procedure was repeated for each dictionary-based technique and for each of the six modes of preprocessing. The complete list of optimal thresholds is provided in Appendix 5, but in most cases, the best performance was achieved, with the threshold of at least 0.5% of all unique terms in the document being present in the respective dictionaries.

### *Data preprocessing*

We lowercased all words in the training and validation datasets to avoid potential inconsistencies and removed punctuation using the Python String package (Python, 2022).

Then, we systematically compared the models' performance in six preprocessing approaches: (1) no additional preprocessing; (2) stopword removal (using German stopwords from the Python Natural Language Toolkit (NLTK) library; Bird et al. 2009); (3) stemming (using the Cistem stemmer from the NLTK library; Bird et al. 2009); (4) lemmatisation (using German lemmatiser from the SpaCy library; Honnibal et al. 2020); (5) stopword removal + stemming; and (6) stopword removal + lemmatisation. Altogether, this process resulted in 66 model combinations (11 models x 6 preprocessing approaches).

*Validation of detection models*

To evaluate the models' performance, we created three validation datasets: (1) a *test validation dataset* (TVD) made up of a subsample of training data (805 journalistic stories; 20% of the training sample); (2) a *diverse validation dataset* (DVD) made of 594 short (e.g. tweets) and long content pieces (e.g. articles from German right-wing outlets); and (3) a *web-tracking validation dataset* (WVD) consisting of 262 documents coming from the corpus of web-tracking data. The TVD was produced following the same principle as the training data (see above), whereas the DVD and WVD were manually annotated (see Appendix 1).

The three datasets were characterised by various degrees of content diversity as well as noise, defined as the "difference in the surface form of an electronic text from the intended, correct or original text" (Agarwal et al., 2007, p. 5). The TVD and DVD had little noise because their content was crawled from a small selection of platforms and carefully parsed. In terms of content diversity, the TVD made only of news articles was the least diverse, whereas the DVD had more diversity. Finally, the WVD had the highest amount of noise, as well as the most content diversity, because its content came from a broad range of platforms to which a platform-agnostic html parser (based on the Selectolax Python library; Golubin 2022) was applied.[5]

To measure the models' performance, we calculated the set of metrics commonly used in the NLP research—precision, recall, and F1 scores—for both predicted classes (i.e. political and non-political), together with the average values. Precision is the ratio of true positive cases to the sum of true positives and false negatives, recall is the ratio of true positives to the sum of true positives and false positives, and the F1 score is the harmonic mean of precision and recall. For readability's sake, we report in the next section only the political class and average F1 scores (for the full metrics, see Appendix 6).

It is important to note that no cross-validation was used when measuring the performance of the models, which made our observations about their performance less robust. This is a major limitation of the study that is attributed to the large number of models compared and the limited computational resources available, with the latter factor being particularly relevant for more computationally demanding techniques (e.g. the ones using BERT). Instead, we opted for a fixed train–test split to make the comparison between the models more consistent by ensuring that all models used the same data for training.

## Findings

Table 1 demonstrates that the SML- (passive-aggressive model) and NN-based models (BERT) achieve the best performance on the TVD, with F1 scores for the political class reaching 0.89 (PA). Such a performance can be attributed to the TVD being the least noisy dataset, as well as the most similar to the data on which the SML- and NN-based models were trained. The lower performance of the CNN model can be attributed to it being substantially simpler (and less computationally demanding) compared with LSTM and BERT. The dictionary-based models, in particular Di-CAP, showed acceptable results but performed worse than SML- and NN-based models that is observations that aligns with earlier comparative NLP studies (e.g. Van Atteveldt et al. 2021).

[5] The necessity to rely on platform-agnostic html parsers is a major challenge associated with the use of web-tracking data (i.e. our use case), as well as the related problem of platform-agnostic automated content analysis. Their use results in the higher volume of noise associated with parsing errors (e.g. infection of organic text with malparsed html tags), but the only alternative is the use of platform-specific platforms, which is not feasible when the data come from thousands of platforms.

**Table 1**. Models' performance on the test validation dataset (TVD)[6]

| | No preprocessing | Stopword removal | Stemming | Stemming + stopword removal | Lemmatisation | Lemmatisation + stopword removal |
|---|---|---|---|---|---|---|
| Di-CAP | 0.76 [0.79] | 0.75 [0.77] | 0.69 [0.71] | 0.69 [0.70] | 0.76 [0.78] | 0.75 [0.77] |
| Di-LL | 0.71 [0.71] | 0.70 [0.70] | 0.67 [0.67] | 0.72 [0.74] | 0.66 [0.69] | 0.67 [0.69] |
| Di-CAP-LL | 0.76 [0.77] | 0.72 [0.73] | 0.71 [0.71] | 0.73 [0.75] | 0.74 [0.77] | 0.74 [0.75] |
| SML [PA] | **0.88 [0.91]** | **0.89 [0.91]** | **0.88 [0.90]** | **0.88 [0.91]** | **0.89 [0.91]** | **0.89 [0.91]** |
| SML [BNB] | 0.86 [0.89] | 0.86 [0.89] | 0.85 [0.88] | 0.86 [0.88] | 0.88 [0.90] | 0.88 [0.90] |
| SML [MNB] | 0.88 [0.90] | **0.89 [0.91]** | 0.87 [0.89] | 0.88 [0.90] | 0.89 [0.91] | 0.88 [0.90] |
| SML [LR] | 0.87 [0.89] | 0.86 [0.88] | 0.86 [0.89] | 0.87 [0.89] | 0.85 [0.88] | 0.87 [0.89] |
| SML [SGD] | 0.83 [0.86] | 0.86 [0.89] | 0.86 [0.89] | 0.85 [0.88] | 0.86 [0.89] | 0.85 [0.88] |
| NN [CNN] | 0.73 [0.80] | 0.82 [0.86] | 0.49 [0.58] | 0.83 [0.87] | 0.82 [0.86] | 0.83 [0.86] |
| NN [LSTM] | 0.79 [0.83] | 0.87 [0.90] | 0.85 [0.88] | 0.79 [0.83] | 0.86 [0.88] | 0.85 [0.88] |
| NN [BERT] | 0.86 [0.88] | 0.81 [0.82] | 0.81 [0.83] | 0.79 [0.79] | 0.85 [0.87] | 0.83 [0.84] |

In terms of preprocessing, we observed close to no impact on the best-performing SML-based models (i.e. a change in the range of 0.01–0.02 for the F1 scores) and little impact on the best-performing NN-based models (i.e. changes in the range of 0.02–0.06). While lemmatisation provided one of the best results, similar scores were also achieved with only stopword removal or no processing (e.g. for BERT). In the case of dictionaries, we observed a similar pattern with lemmatisation and no processing, which provided optimal results.

[6] In this and the following tables, the highest performance values per preprocessing mode are bolded.

For the DVD (Table 2), the best performance was achieved by BERT (up to a 0.90 F1 score for the political class). The other NN- and SML-based models showed a major drop in performance, which can be attributed to the increase in content diversity. We expect that additional fine-tuning (e.g. complications in the network architecture or an increase in the number of learning epochs) would improve these models' performance; however, under the condition of the non-fine-tuned low-cost implementation, BERT provided substantially better results. The second best performance (0.77 F1 score for the political class) was achieved by Di-CAP-LL, which can be attributed to dictionary-based models being more robust when dealing with diverse content.

**Table 2**. Models' performance on the diverse validation dataset (DVD)

| | No preprocessing | Stopword removal | Stemming | Stemming + stopword removal | Lemmatisation | Lemmatisation + stopword removal |
|---|---|---|---|---|---|---|
| Di-CAP | 0.75 [0.70] | 0.75 [0.70] | 0.75 [0.65] | 0.69 [0.63] | 0.75 [0.7] | 0.76 [0.70] |
| Di-LL | 0.66 [0.58] | 0.66 [0.57] | 0.64 [0.55] | 0.63 [0.58] | 0.56 [0.52] | 0.56 [0.51] |
| Di-CAP-LL | 0.75 [0.68] | 0.76 [0.66] | 0.77 [0.67] | 0.73 [0.66] | 0.73 [0.66] | 0.56 [0.51] |
| SML [PA] | 0.49 [0.51] | 0.47 [0.50] | 0.52 [0.53] | 0.50 [0.51] | 0.50 [0.51] | 0.49 [0.51] |
| SML [BNB] | 0.36 [0.43] | 0.36 [0.43] | 0.38 [0.44] | 0.44 [0.48] | 0.38 [0.44] | 0.43 [0.47] |
| SML [MNB] | 0.69 [0.65] | 0.69 [0.65] | 0.63 [0.60] | 0.68 [0.64] | 0.67 [0.63] | 0.71 [0.67] |
| SML [LR] | 0.44 [0.48] | 0.41 [0.46] | 0.46 [0.49] | 0.43 [0.48] | 0.44 [0.48] | 0.42 [0.47] |
| SML [SGD] | 0.65 [0.61] | 0.37 [0.43] | 0.59 [0.57] | 0.62 [0.58] | 0.58 [0.56] | 0.66 [0.61] |
| NN [CNN] | 0.15 [0.30] | 0.27 [0.37] | 0.29 [0.34] | 0.36 [0.43] | 0.32 [0.41] | 0.35 [0.42] |
| NN [LSTM] | 0.53 [0.53] | 0.44 [0.48] | 0.54 [0.55] | 0.54 [0.54] | 0.52 [0.53] | 0.53 [0.53] |
| NN [BERT] | **0.85 [0.78]** | **0.90 [0.83]** | **0.87 [0.79]** | **0.89 [0.81]** | **0.86 [0.79]** | **0.90 [0.84]** |

Similar to the TVB, the impact of preprocessing was limited for the DVD. For BERT, we observed changes in the range of 0.01–0.05, with the best results achieved by stopword removal and lemmatisation with stopword removal. A similar effect was observed for the dictionary-based models, where stopword removal or lemmatisation/stemming enabled the best performance. The largest preprocessing-based fluctuations were observed for CNN, where the addition of more complex forms of preprocessing led to a major increase (i.e. up to 0.21) in the F1 score for the political class; however, the overall performance of the model remained too low to be usable.

Finally, for the noisiest and most diverse validation dataset (WVD; Table 3), we observed the best performance from the dictionary-based models. The F1 scores for these models reached 0.81 (Di-CAP) and 0.78 (Di-LL/Di-CAP-LL) for the political class, as contrasted to the 0.63 F1 score for the best NN-based model (BERT). Similar to the TVD, the addition of the log-likelihood sub-dictionary usually worsened the performance of the CAP-based sub-dictionary (except for the case when the Di-CAP-LL was combined with stemming); the latter observation can be explained by the Di-LL bringing additional semantic noise that is more detrimental for highly diverse data.

**Table 3**. Models' performance on the web-tracking validation dataset (WVD)

| | No preprocessing | Stopword removal | Stemming | Stemming + stopword removal | Lemmatisation | Lemmatisation + stopword removal |
|---|---|---|---|---|---|---|
| Di-CAP | **0.79 [0.83]** | **0.79 [0.83]** | 0.62 [0.72] | **0.72 [0.77]** | **0.81 [0.85]** | **0.81 [0.85]** |
| Di-LL | 0.68 [0.75] | 0.65 [0.73] | 0.59 [0.66] | 0.61 [0.66] | 0.63 [0.70] | 0.63 [0.70] |
| Di-CAP-LL | 0.73 [0.79] | 0.76 [0.81] | **0.63 [0.72]** | 0.67 [0.75] | 0.78 [0.83] | 0.78 [0.82] |
| SML [PA] | 0.47 [0.65] | 0.44 [0.63] | 0.48 [0.65] | 0.42 [0.62] | 0.45 [0.63] | 0.42 [0.61] |
| SML [BNB] | 0.22 [0.51] | 0.30 [0.55] | 0.28 [0.54] | 0.23 [0.51] | 0.22 [0.51] | 0.22 [0.51] |
| SML [MNB] | 0.44 [0.63] | 0.44 [0.63] | 0.34 [0.57] | 0.46 [0.64] | 0.39 [0.60] | 0.43 [0.62] |
| SML [LR] | 0.50 [0.67] | 0.49 [0.66] | 0.50 [0.67] | 0.49 [0.66] | 0.52 [0.68] | 0.48 [0.66] |
| SML [SGD] | 0.48 [0.63] | 0.47 [0.65] | 0.41 [0.59] | 0.38 [0.57] | 0.51 [0.66] | 0.52 [0.66] |
| NN [CNN] | 0.12 [0.45] | 0.28 [0.54] | 0.30 [0.51] | 0.38 [0.59] | 0.34 [0.57] | 0.46 [0.64] |
| NN [LSTM] | 0.1 [0.54] | 0.36 [0.58] | 0.37 [0.59] | 0.46 [0.61] | 0.37 [0.58] | 0.24 [0.51] |
| NN [BERT] | 0.32 [0.56] | 0.58 [0.71] | 0.54 [0.67] | 0.63 [0.71] | 0.45 [0.63] | 0.63 [0.72] |

From the preprocessing point of view, lemmatisation again delivered the best results, followed by stopword removal and the absence of preprocessing. The small variation between the best-performing scores (i.e. 0.02) suggests that the use of more computationally demanding options might not be justified, especially when dealing with the large volumes of cross-platform data. Interestingly, stemming led to a substantial performance drop (0.62 compared with 0.79 with no preprocessing for Di-CAP), which may be due to its creation of artificial noise (e.g. by creating unwanted ambiguities caused by the resolution of words to their stems).

**Discussion**

The increase in the volume and diversity of political content available online prompts the need for platform-agnostic approaches for its detection. In this paper, we discuss various ways to address this problem by comparing three groups of detection techniques—dictionary, SML, and NN-based ones—and examining the impact of preprocessing on their performance.

Our results show that while the SML- and NN-based models demonstrated solid performance on content similar to the type they were trained on (i.e. journalistic articles in our case), their effectiveness dropped for more diverse content. The major exception here is BERT, which highlights the ability of more computationally demanding transformer models to achieve high performance on data coming from a broad range of platforms; this finding aligns with the existing evaluations of transformers being the state-of-the-art approach in the field of NLP (e.g. Biggiogera et al. 2021). However, for web-tracking data with a high amount of noise, even BERT's performance turned out to be low.

While dictionaries did not show the best performance on less noisy data, for web-tracking data, they outperformed more complex techniques. There may be two reasons for this. First, the underlying principle of the dictionary (i.e. word matching) might be more fitting for detection tasks with a specific focus (e.g. phenomena associated with a concrete set of actors) and less vulnerable to data noise (e.g. html artefacts left after parsing). Second, the high scores of Di-CAP (i.e. dictionary combining actor/institution names and CAP terms) can be attributed to it being manually verified (and, hence, more fitting for the task) than journalistic tag-based datasets used for training other techniques. The latter interpretation is supported by the drop in the performance of the combined dictionary (i.e., Di-LL-CAP), which may be due to the LL dictionary bringing additional noise. This difference highlights the fact that the combination of dictionaries does not automatically lead to performance improvement.

The high performance of Di-CAP does not mean that dictionary-based techniques always outperform SML- and NN-based techniques. However, under the condition of limited development resources, the reuse of an existing dictionary with the possibility of additional augmentations might provide better results than reliance on metadata-based labels (e.g. journalistic tags) for SMLs and NNs. With enough resources available, we expect a diverse corpus of manually annotated training data to potentially outperform dictionaries, as has been the case with other automated content analysis tasks (Van Atteveldt et al. 2021).

These results have several implications for the use of automated content analysis in political communication research. First, they show that designing platform-agnostic detection approaches is possible, even though the process of doing so remains rather challenging. The success of such an endeavour depends on the robustness of the chosen approach and the amount of available resources (either in the form of in-house development capacities or third-party assets made accessible by the research community). The combination of these two factors makes dictionary-based approaches particularly appealing: not only might they be less subjected to data noise (especially in the case of cross-platform forms of passive data collection; Stier et al. 2020), but the procedure for asset reuse is also more intuitive for dictionaries compared with SML- or NN-based models.

Second, our results resonate with earlier calls (e.g. Grimmer and Stewart 2013) for the extensive validation of automated content analysis techniques. While some SML and NN models showed high performance on the TVD, it substantially worsened on less familiar and more diverse content and dropped even further on data containing a high volume of noise. These observations stress the importance of utilising more than one validation dataset for measuring the performance of the NLP techniques used to study communication phenomena, together with making these validation datasets diverse (i.e. by including content coming from different platforms), especially when aiming to make these techniques platform-agnostic.

Finally, our study offers insights into the impact of preprocessing on different NLP techniques. Specifically, it suggests that there is little difference between the models' performance under the conditions of no preprocessing and when more complex preprocessing modes are used. This observation is particularly relevant for more computationally demanding modes (e.g. lemmatisation), where marginal increases in performance do not necessarily justify the time and resource costs needed to apply them for large datasets. This leads us to a conclusion similar to those of earlier NLP studies (e.g.

HaCohen-Kerner et al. 2020), in that stopword removal is often an optimal mode of preprocessing because it allows for a performance increase at a relatively small cost.

This paper is not without limitations. First, we focused on low-cost implementations of the detection techniques without using additional fine-tuning or training resources (e.g. increases in the number of epochs for NNs). Such a focus is attributed to the expectation that many academic-based projects would have limited resources for implementing such techniques, but it should be taken into consideration that a comparison of high-cost implementations might change the results. Second, the training data for the SML- and NN-based techniques (as well as the Di-LL dictionary) were made exclusively of journalistic articles. While this choice allows for the utilisation of metadata-based labels and, thus, avoids the need for manual labelling of the training data, it results in the bias of models towards specific formats of content (i.e. journalistic articles) and makes it harder for models other than BERT to deal with more diverse and noisy data. Future research could benefit from relying on more diverse sets of training data, especially when dealing with platform-agnostic detection tasks. Third, because of the large number of models compared, together with the limited computational resources, we did not use cross-validation to make the evaluations of our models' performance more robust. For future research, it is important to integrate cross-validation into the process of performance evaluation to obtain more generalisable insights.

# Appendix

## Appendix A1: Definition of politics-related content and procedure for manually labelling it for validation datasets

Our understanding of politics-related information is based on three dimensions that include (1) processes and political procedures (politics); (2) form, structures, and institutional aspects (polity); and (3) the content of political disputes (policy; see, e.g., Nohlen and Thibaut 2011). When developing the codebook to label the content used to create the validation datasets for our detection approaches, we assumed the piece of content contained politics-related information if the respective piece of content related to at least one of the above-mentioned dimensions.

To prepare the two validation datasets the classification of political content (i.e. DVD and WVD; the labelling of the last validation dataset [TVD] was based on the metadata in the form of journalistic tags), the coders manually labelled two sets of documents (for a summary of this process, see Table A1.1 below). The first set consisted of 594 documents made of short (e.g. Twitter and Telegram posts by politicians and journalists) and longer documents (articles from German legacy media, such as *Süddeutsche Zeitung*, and right-wing outlets, like *Journalistenwatch*) crawled online. The second set consisted of 262 documents randomly sampled from the web-tracking data.

**Table A1.1**. Summary of the procedures for preparing the validation datasets

| Dataset name | Size | Source | Preparation | Coding validation |
|---|---|---|---|---|
| Test validation dataset (TVD) | 805 documents | Journalistic articles from *Blick*, *Bild*, and *Süddeutsche Zeitung* | No manual labelling; journalistic metadata used as labels | No validation |
| Diverse validation dataset (DVD) | 594 documents | Short (Twitter and Telegram posts by politicians and journalists) and longer documents (articles from German legacy media, such as *Süddeutsche* Zeitung, and right-wing outlets, like *Journalistenwatch*) crawled online | 4 coders | 10% of the dataset used for reliability check: Cohen's Kappa = .86 (political actor) and Cohen's Kappa = .68 (political topic); complete dataset checked and consensus-coded by 2 experts |
| Web-tracking validation dataset (WVD) | 262 documents | Random sample of web-tracking data | 1 coder | For all cases where the manual coding was inconsistent with the classification, the coding was reviewed, discussed, and consensus-coded by 3 experts |

Each document in the two datasets (DVD and WVD) was coded according to the following three variables: (1) POLITICAL ACTOR, POLITICAL INSTITUTIONS, AND PRINCIPLES (i.e. *Are political actors mentioned in the text? Does the text mention political institutions?*); (2) POLITICAL TOPICS (i.e. *Are political topics/issues mentioned in the text?*); and (3) OTHER POLITICAL SUBJECT AREA. A detailed description of the variables is provided below; the document was classified as political if at least one of these variables was present in the document.

**Variable: POLITICAL ACTOR, POLITICAL INSTITUTIONS, AND PRINCIPLES**

***Are political actors mentioned in the text? Does the text mention political institutions?***

0 = no

1 = yes

*Municipal/regional/national international political actors such as:*

- (Members of) governments (e.g. government, president, chancellor, or minister)
- Opposition (e.g. opposition leader or opposition parliamentary group)
- Parliamentarian (e.g. member of the Bundestag or Nationalrat)
- Politician
- Party (e.g. party leader or members)

- Administration (e.g. ministry)

*Politically active citizens who take clear political action such as:*

- Lobbying specific issues
- Formulating demands placed on politicians
- Organising demonstrations

*Entire state as political actors (i.e. local/regional/national/international political institutions and principles) such as:*

- Constitution
- Political institutions (e.g. parliament, Council of Europe, European Union, or courts of law)
- Laws and agreements
- Political culture (e.g. political attitudes of citizens)

**Variable: POLITICAL TOPICS**

***Are political topics/issues mentioned in the text?***

0 = no

1 = yes

*Political topics/issues at the municipal/regional/national/international level:*

- Working conditions and labour market
- Foreign and domestic trade
- Banks and finances
- Education
- Civil rights and liberties and minority rights
- German reunification

- Energy
- Family and social affairs
- Health care
- International relations and foreign aid
- Agriculture
- Mobility, transport, and traffic
- Public administration
- Political campaigns, referendums, initiatives, and votes
- Law and crime
- Environment
- Defence
- Water management
- Housing, construction, and spatial planning
- Economy
- Science and technology
- Other subject area if applicable: the subject is to be named in the variable ‘OTHER POLITICAL SUBJECT AREA’

A. Content related to these subject areas was considered political if it presented problems/proposed solutions from a perspective relevant to a population group or society as a whole (not just to an individual).

B. Political issues could be described at different stages of the process:

- As a pure description of the problem (without a policy demand or action)
- As a demand for policy (still without action)

- As political discussions or actions

**Variable: OTHER POLITICAL SUBJECT AREA**

*Information deals with political content in a subject area that was not mentioned in the variable 'POLITICAL TOPICS':*

- Naming of the subject area

## Appendix A2: Use case

Our interest in developing platform-agnostic approaches for detecting politics-related information is related to the specific requirements of our use case that deals with web-tracking data. Similar to other web-tracking studies relying on a large number of participants (e.g. Stier et al. 2020), we acquired data from a sample of participants (N = 1,149) from Germany and Switzerland who were recruited via a market research company and subsequently installed web-tracking software on their desktop browsers. During the two web-tracking rounds in spring and autumn 2020, the software extracted the HTML content appearing in the browsers and then transferred it to the remote server by following the screen-scraping principle (for more information, see Christner et al. 2021).

To protect the participants' privacy, we used a denylist (i.e. a list of websites visits to that were not captured, which in this case included healthcare-related websites, online banking portals, commercial websites, and pornography), as well as customised filters for social media platforms (i.e. to avoid capturing personally identifiable data as well as private data). The remaining visits were all captured, resulting in over four million html pages recorded from almost 90,000 unique domains. Such diversity makes it infeasible to develop platform-specific detection solutions capable of accounting for the distinct format and style of textual data associated with a specific platform. Hence, to be able to detect politics-related content in these multi-platform data, as well as to realise possibilities provided by these data (e.g. by looking at how the long tail of information consumption affects individual engagements with politics-related information), we needed a platform-agnostic means of automated detection.

## Appendix A3: NN-based model architectures

To train the CNN- and LSTM-based models, we used tokenisers with the 120,000 most common words in the training data. We used a batch size of 256 with five training epochs for each model, with a binary cross entropy function for measuring the loss during the model training, and a Nadam optimiser. For both the CNN and LSTM models, we used sequential architectures with the composition of layers described in Tables A3.1 and A3.2.

**Table A3.1**. Composition of layers for the CNN network

| Layer | Activation |
|---|---|
| Embedding | None |
| 1D convolution layer | ReLU |
| Global max pooling for 1D data | None |
| Dropout layer | None |
| Dense layer | ReLU |
| Dropout layer | None |
| Dense layer | Sigmoid |

**Table A3.2**. Composition of layers for the LSTM network

| Layer | Activation |
|---|---|
| Embedding | None |
| Bidirectional LSTM | None |
| Global max pooling for 1D data | None |
| Dropout layer | None |
| Dense layer | ReLU |
| Dropout layer | None |
| Dense layer | Sigmoid |

## Appendix A4: Probability of assigning the "political" label by BERT models

BERT allows for the specification of the threshold of the minimal probability of a specific label (e.g. "political") being assigned to the document to which the model is applied. The lower this probability is, the less conservative is the prediction, and vice versa. To identify the optimal probability, we tested different thresholds (i.e. 5%, 10%, 15%, 20%, 25%, 30%, 35%, and 40%) for the three validation datasets using the average F1 score as a performance measurement. The results of the testing process are provided in Table A4.1; based on the comparison of the aggregated F1 scores across all three datasets, we opted for a 15% probability that showed the highest performance.

**Table A4.1.** Performance of BERT models per probability threshold (average F1 scores)

| | Probability / preprocessing | 0.05 | 0.1 | 0.15 | 0.2 | 0.25 | 0.3 | 0.35 | 0.4 |
|---|---|---|---|---|---|---|---|---|---|
| TVD | No-pre | 0.82 | 0.86 | 0.88 | 0.88 | 0.88 | 0.89 | 0.88 | 0.88 |
| | Stop | 0.74 | 0.79 | 0.82 | 0.84 | 0.85 | 0.87 | 0.88 | 0.89 |
| | Ste | 0.73 | 0.80 | 0.83 | 0.85 | 0.86 | 0.87 | 0.87 | 0.88 |
| | Ste+stop | 0.69 | 0.76 | 0.79 | 0.82 | 0.83 | 0.85 | 0.86 | 0.87 |
| | Lem | 0.80 | 0.84 | 0.87 | 0.88 | 0.88 | 0.89 | 0.89 | 0.90 |
| | Lem+stop | 0.75 | 0.80 | 0.84 | 0.85 | 0.87 | 0.88 | 0.88 | 0.88 |
| DVD | No-pre | 0.82 | 0.80 | 0.78 | 0.76 | 0.75 | 0.72 | 0.70 | 0.68 |
| | Stop | 0.81 | 0.82 | 0.83 | 0.81 | 0.80 | 0.79 | 0.78 | 0.77 |
| | Ste | 0.78 | 0.80 | 0.79 | 0.78 | 0.76 | 0.75 | 0.73 | 0.71 |
| | Ste+stop | 0.80 | 0.82 | 0.81 | 0.81 | 0.79 | 0.78 | 0.76 | 0.74 |
| | Lem | 0.82 | 0.81 | 0.79 | 0.77 | 0.76 | 0.74 | 0.72 | 0.71 |
| | Lem+stop | 0.85 | 0.85 | 0.84 | 0.83 | 0.83 | 0.82 | 0.80 | 0.79 |
| WVD | No-pre | 0.64 | 0.58 | 0.56 | 0.55 | 0.54 | 0.52 | 0.50 | 0.48 |

| | | | | | | | | |
|---|---|---|---|---|---|---|---|---|
| Stop | 0.70 | 0.72 | 0.71 | 0.68 | 0.67 | 0.66 | 0.62 | 0.61 |
| Ste | 0.66 | 0.71 | 0.67 | 0.65 | 0.63 | 0.63 | 0.59 | 0.57 |
| Ste+stop | 0.64 | 0.70 | 0.71 | 0.71 | 0.70 | 0.69 | 0.66 | 0.64 |
| Lem | 0.65 | 0.64 | 0.63 | 0.61 | 0.58 | 0.57 | 0.55 | 0.55 |
| Lem+stop | 0.67 | 0.70 | 0.72 | 0.73 | 0.72 | 0.71 | 0.67 | 0.68 |

## Appendix A5: Thresholds for dictionary-based techniques

In the case of dictionary-based techniques, the decision of whether the document contained information related to politics was based on the presence of politics-related terms within a document. To address the possible variation in the length of documents coming from different platforms, we relied not on the absolute numbers but on the ratio between the number of politics-related unique terms to the overall number of unique terms per document. Table A5.1 shows the optimal ratios of unique terms from the individual dictionaries to all unique terms per document for each dictionary-based technique depending on a specific preprocessing mode. The document was to be classified as political if the ratio of unique political to non-political terms was equal to or exceeded the optimal ratio. For instance, in the case of the Di-CAP dictionary with no preprocessing mode, at least 0.48% of all unique terms present in the document had to be in the political dictionary for the document to be classified as political.

**Table A5.1**. Optimal thresholds for dictionary-based techniques for detecting politics-related information

| Preprocessing mode | Di-CAP | Di-LL | Di-CAP-LL |
|---|---|---|---|
| No-preprocessing | 0.0048 | 0.0077 | 0.0131 |
| Stopword removal | 0.0041 | 0.0093 | 0.0138 |
| Stemming | 0.0080 | 0.0056 | 0.0140 |
| Stemming+stopword removal | 0.0052 | 0.0049 | 0.0117 |
| Lemmatisation | 0.0048 | 0.0043 | 0.0089 |
| Lemmatisation+stopword removal | 0.0049 | 0.0050 | 0.0094 |

**Appendix A6: Performance metrics for politics-related information detection models**

This appendix provides a detailed overview of the performance metrics for the politics-related information detection models. Specifically, it provides information about the model performance for non-political (i.e. 0/non-pol class) and political (i.e. 1/pol class) detection together with the average performance values. Three metrics are provided: (1) *precision* (pr), which shows how many of the cases identified by the model as positive were actually positive, is calculated by dividing the number of true positives by the sum of true positives and false negatives; (2) *recall* (rec), which shows how many of all the positive cases available were correctly identified by the model, is calculated by dividing the true positive rate by the sum of true positives and false positives; and (3) F1 score (F1), which is the harmonic mean of precision and recall.

The metrics are provided in Tables A6.1 to A6.6. Each table corresponds to one of the six modes of preprocessing (e.g. no preprocessing or stemming only) and contains information about the performance of the models across the following three validation datasets: (1) a *test validation dataset* (TVD) made of a subsample of training data (805 journalistic stories; 20% of the training sample); (2) a *diverse validation dataset* (DVD) comprising 594 short (e.g. tweets) and long content pieces (e.g. articles from German right-wing validation outlets); and (3) a *web-tracking validation dataset* (WVD) consisting of 262 documents from the corpus of web-tracking data.

**Table A6.1.** Comparison of techniques for detecting politics-related information (lowercasing/no preprocessing)

| | TVD | | | DVD | | | WVD | | |
|---|---|---|---|---|---|---|---|---|---|
| | Pr | Rec | F1 | Pr | Rec | F1 | Pr | Rec | F1 |
| *Di-CAP [av]* | 0.79 | 0.81 | 0.79 | 0.72 | 0.77 | 0.68 | 0.86 | 0.84 | 0.85 |
| 0 [non-pol] | 0.90 | 0.75 | 0.82 | 0.47 | 0.96 | 0.63 | 0.87 | 0.92 | 0.89 |
| 1 [pol] | 0.68 | 0.86 | 0.76 | 0.97 | 0.58 | 0.72 | 0.85 | 0.77 | 0.81 |
| *Di-LL [av]* | 0.75 | 0.75 | 0.71 | 0.60 | 0.62 | 0.58 | 0.75 | 0.75 | 0.75 |
| 0 [non-pol] | 0.92 | 0.59 | 0.72 | 0.38 | 0.69 | 0.49 | 0.81 | 0.81 | 0.81 |
| 1 [pol] | 0.58 | 0.92 | 0.71 | 0.60 | 0.62 | 0.58 | 0.68 | 0.68 | 0.68 |
| *Di-CAP-LL [av]* | 0.79 | 0.80 | 0.77 | 0.69 | 0.74 | 0.68 | 0.81 | 0.79 | 0.79 |
| 0 [non-pol] | 0.93 | 0.68 | 0.79 | 0.48 | 0.83 | 0.61 | 0.82 | 0.89 | 0.86 |
| 1 [pol] | 0.64 | 0.92 | 0.76 | 0.91 | 0.65 | 0.75 | 0.79 | 0.68 | 0.73 |
| *SML [PA] [av]* | 0.91 | 0.9 | 0.91 | 0.68 | 0.66 | 0.51 | 0.81 | 0.65 | 0.65 |
| 0 [non-pol] | 0.91 | 0.95 | 0.93 | 0.36 | 0.99 | 0.53 | 0.71 | 0.98 | 0.82 |
| 1 [pol] | 0.91 | 0.85 | 0.88 | 0.99 | 0.33 | 0.49 | 0.91 | 0.32 | 0.47 |
| *SML [BNB] [av]* | 0.9 | 0.89 | 0.89 | 0.67 | 0.61 | 0.43 | 0.83 | 0.56 | 0.51 |
| 0 [non-pol] | 0.91 | 0.93 | 0.92 | 0.33 | 1 | 0.5 | 0.66 | 1.00 | 0.79 |
| 1 [pol] | 0.88 | 0.85 | 0.86 | 1 | 0.22 | 0.36 | 1.00 | 0.12 | 0.22 |
| *SML [MNB] [av]* | 0.9 | 0.91 | 0.9 | 0.71 | 0.75 | 0.65 | 0.83 | 0.64 | 0.63 |
| 0 [non-pol] | 0.95 | 0.9 | 0.92 | 0.45 | 0.98 | 0.61 | 0.70 | 0.99 | 0.82 |
| 1 [pol] | 0.85 | 0.92 | 0.88 | 0.98 | 0.53 | 0.69 | 0.97 | 0.29 | 0.44 |
| *SML [LR] [av]* | 0.89 | 0.89 | 0.89 | 0.67 | 0.64 | 0.48 | 0.86 | 0.67 | 0.67 |
| 0 [non-pol] | 0.92 | 0.91 | 0.92 | 0.35 | 0.99 | 0.51 | 0.72 | 1.00 | 0.83 |
| 1 [pol] | 0.85 | 0.88 | 0.87 | 0.99 | 0.28 | 0.44 | 1.00 | 0.34 | 0.50 |
| *SML [SGD] [av]* | 0.86 | 0.86 | 0.86 | 0.68 | 0.72 | 0.61 | 0.66 | 0.62 | 0.63 |
| 0 [non-pol] | 0.91 | 0.87 | 0.89 | 0.42 | 0.93 | 0.58 | 0.70 | 0.86 | 0.77 |
| 1 [pol] | 0.8 | 0.86 | 0.83 | 0.95 | 0.5 | 0.65 | 0.62 | 0.39 | 0.48 |
| *NN [CNN] [av]* | 0.86 | 0.78 | 0.80 | 0.65 | 0.54 | 0.30 | 0.82 | 0.53 | 0.45 |
| 0 [non-pol] | 0.80 | 0.97 | 0.88 | 0.30 | 1.00 | 0.46 | 0.64 | 1.00 | 0.78 |
| 1 [pol] | 0.92 | 0.60 | 0.73 | 1.00 | 0.08 | 0.15 | 1.00 | 0.06 | 0.12 |
| *NN [LSTM] [av]* | 0.82 | 0.83 | 0.83 | 0.65 | 0.65 | 0.53 | 0.62 | 0.56 | 0.54 |
| 0 [non-pol] | 0.89 | 0.84 | 0.86 | 0.37 | 0.93 | 0.53 | 0.66 | 0.91 | 0.76 |
| 1 [pol] | 0.76 | 0.83 | 0.79 | 0.93 | 0.37 | 0.53 | 0.58 | 0.21 | 0.31 |
| *NN [BERT] [av]* | 0.87 | 0.89 | 0.88 | 0.77 | 0.82 | 0.78 | 0.72 | 0.58 | 0.56 |
| 0 [non-pol] | 0.95 | 0.84 | 0.90 | 0.60 | 0.86 | 0.71 | 0.67 | 0.96 | 0.79 |
| 1 [pol] | 0.79 | 0.93 | 0.86 | 0.93 | 0.77 | 0.85 | 0.77 | 0.20 | 0.32 |

**Table A6.2**. Comparison of techniques for detecting politics-related information (lowercasing/no stopwords)

| | TVD | | | DVD | | | WVD | | |
|---|---|---|---|---|---|---|---|---|---|
| | Pr | Rec | F1 | Pr | Rec | F1 | Pr | Rec | F1 |
| *Di-CAP [av]* | 0.78 | 0.80 | 0.77 | 0.73 | 0.78 | 0.70 | 0.82 | 0.83 | 0.83 |
| 0 [non-pol] | 0.92 | 0.69 | 0.79 | 0.49 | 0.95 | 0.64 | 0.88 | 0.85 | 0.87 |
| 1 [pol] | 0.64 | 0.90 | 0.75 | 0.97 | 0.61 | 0.75 | 0.77 | 0.81 | 0.79 |
| *Di-LL [av]* | 0.75 | 0.75 | 0.70 | 0.58 | 0.60 | 0.57 | 0.73 | 0.72 | 0.73 |
| 0 [non-pol] | 0.93 | 0.56 | 0.70 | 0.37 | 0.64 | 0.47 | 0.79 | 0.82 | 0.80 |
| 1 [pol] | 0.56 | 0.93 | 0.70 | 0.80 | 0.57 | 0.66 | 0.67 | 0.63 | 0.65 |
| *Di-CAP-LL [av]* | 0.76 | 0.77 | 0.73 | 0.66 | 0.70 | 0.66 | 0.81 | 0.81 | 0.81 |
| 0 [non-pol] | 0.94 | 0.60 | 0.73 | 0.47 | 0.72 | 0.56 | 0.85 | 0.87 | 0.86 |
| 1 [pol] | 0.59 | 0.94 | 0.72 | 0.86 | 0.68 | 0.76 | 0.77 | 0.74 | 0.76 |
| *SML [PA] [av]* | 0.92 | 0.91 | 0.91 | 0.68 | 0.65 | 0.5 | 0.83 | 0.64 | 0.63 |
| 0 [non-pol] | 0.92 | 0.95 | 0.93 | 0.36 | 0.99 | 0.53 | 0.70 | 0.99 | 0.82 |
| 1 [pol] | 0.91 | 0.87 | 0.89 | 0.99 | 0.31 | 0.47 | 0.97 | 0.29 | 0.44 |
| *SML [BNB] [av]* | 0.9 | 0.89 | 0.89 | 0.67 | 0.61 | 0.43 | 0.83 | 0.59 | 0.55 |
| 0 [non-pol] | 0.91 | 0.93 | 0.92 | 0.33 | 1 | 0.5 | 0.67 | 1.00 | 0.80 |
| 1 [pol] | 0.88 | 0.85 | 0.86 | 1 | 0.22 | 0.36 | 1.00 | 0.17 | 0.30 |
| *SML [MNB] [av]* | 0.9 | 0.91 | 0.91 | 0.71 | 0.75 | 0.65 | 0.82 | 0.64 | 0.63 |
| 0 [non-pol] | 0.95 | 0.9 | 0.93 | 0.44 | 0.98 | 0.61 | 0.70 | 0.99 | 0.82 |
| 1 [pol] | 0.85 | 0.93 | 0.89 | 0.98 | 0.53 | 0.69 | 0.93 | 0.29 | 0.44 |
| *SML [LR] [av]* | 0.88 | 0.88 | 0.88 | 0.67 | 0.63 | 0.46 | 0.86 | 0.66 | 0.66 |
| 0 [non-pol] | 0.91 | 0.91 | 0.91 | 0.34 | 1 | 0.51 | 0.71 | 1.00 | 0.83 |
| 1 [pol] | 0.85 | 0.86 | 0.86 | 1 | 0.26 | 0.41 | 1.00 | 0.33 | 0.49 |
| *SML [SGD] [av]* | 0.89 | 0.88 | 0.89 | 0.67 | 0.61 | 0.43 | 0.85 | 0.65 | 0.65 |
| 0 [non-pol] | 0.91 | 0.92 | 0.91 | 0.33 | 1 | 0.5 | 0.71 | 1.00 | 0.83 |
| 1 [pol] | 0.87 | 0.85 | 0.86 | 1 | 0.22 | 0.37 | 1.00 | 0.31 | 0.47 |
| *NN [CNN] [av]* | 0.89 | 0.85 | 0.86 | 0.66 | 0.58 | 0.38 | 0.80 | 0.58 | 0.54 |
| 0 [non-pol] | 0.86 | 0.96 | 0.91 | 0.32 | 1.00 | 0.48 | 0.67 | 0.99 | 0.80 |
| 1 [pol] | 0.92 | 0.74 | 0.82 | 1.00 | 0.16 | 0.27 | 0.94 | 0.16 | 0.28 |
| *NN [LSTM] [av]* | 0.91 | 0.89 | 0.90 | 0.67 | 0.63 | 0.48 | 0.78 | 0.60 | 0.58 |
| 0 [non-pol] | 0.91 | 0.95 | 0.93 | 0.35 | 0.99 | 0.52 | 0.68 | 0.98 | 0.80 |
| 1 [pol] | 0.91 | 0.84 | 0.87 | 0.98 | 0.28 | 0.44 | 0.88 | 0.22 | 0.36 |
| *NN [BERT] [av]* | 0.84 | 0.85 | 0.82 | 0.82 | 0.84 | 0.83 | 0.77 | 0.70 | 0.71 |
| 0 [non-pol] | 0.98 | 0.73 | 0.84 | 0.72 | 0.80 | 0.76 | 0.74 | 0.93 | 0.83 |
| 1 [pol] | 0.69 | 0.98 | 0.81 | 0.92 | 0.88 | 0.90 | 0.80 | 0.46 | 0.58 |

**Table A6.3**. Comparison of techniques for detecting politics-related information (lowercasing/stemming)

| | TVD | | | DVD | | | WVD | | |
|---|---|---|---|---|---|---|---|---|---|
| | Pr | Rec | F1 | Pr | Rec | F1 | Pr | Rec | F1 |
| *Di-CAP [av]* | 0.72 | 0.74 | 0.71 | 0.65 | 0.69 | 0.65 | 0.74 | 0.71 | 0.72 |
| 0 [non-pol] | 0.87 | 0.62 | 0.72 | 0.45 | 0.72 | 0.56 | 0.76 | 0.88 | 0.82 |
| 1 [pol] | 0.58 | 0.85 | 0.69 | 0.86 | 0.66 | 0.75 | 0.73 | 0.54 | 0.62 |
| *Di-LL [av]* | 0.70 | 0.71 | 0.67 | 0.58 | 0.60 | 0.55 | 0.66 | 0.67 | 0.66 |
| 0 [non-pol] | 0.87 | 0.56 | 0.68 | 0.36 | 0.68 | 0.47 | 0.76 | 0.70 | 0.73 |
| 1 [pol] | 0.54 | 0.86 | 0.67 | 0.81 | 0.52 | 0.64 | 0.56 | 0.63 | 0.59 |
| *Di-CAP-LL [av]* | 0.75 | 0.75 | 0.71 | 0.66 | 0.69 | 0.67 | 0.75 | 0.72 | 0.72 |
| 0 [non-pol] | 0.91 | 0.60 | 0.72 | 0.48 | 0.68 | 0.56 | 0.77 | 0.87 | 0.82 |
| 1 [pol] | 0.58 | 0.91 | 0.71 | 0.85 | 0.71 | 0.77 | 0.72 | 0.56 | 0.63 |
| *SML [PA] [av]* | 0.91 | 0.9 | 0.9 | 0.68 | 0.67 | 0.53 | 0.80 | 0.65 | 0.65 |
| 0 [non-pol] | 0.92 | 0.94 | 0.93 | 0.37 | 0.99 | 0.54 | 0.71 | 0.98 | 0.82 |
| 1 [pol] | 0.89 | 0.86 | 0.88 | 0.99 | 0.36 | 0.52 | 0.89 | 0.33 | 0.48 |
| *SML [BNB] [av]* | 0.88 | 0.88 | 0.88 | 0.67 | 0.62 | 0.44 | 0.83 | 0.58 | 0.54 |
| 0 [non-pol] | 0.91 | 0.9 | 0.91 | 0.33 | 1 | 0.5 | 0.67 | 1.00 | 0.80 |
| 1 [pol] | 0.84 | 0.85 | 0.85 | 1 | 0.23 | 0.38 | 1.00 | 0.16 | 0.28 |
| *SML [MNB] [av]* | 0.89 | 0.9 | 0.89 | 0.69 | 0.71 | 0.6 | 0.84 | 0.6 | 0.57 |
| 0 [non-pol] | 0.94 | 0.9 | 0.92 | 0.41 | 0.96 | 0.57 | 0.68 | 1.00 | 0.81 |
| 1 [pol] | 0.84 | 0.9 | 0.87 | 0.97 | 0.47 | 0.63 | 1.00 | 0.20 | 0.34 |
| *SML [LR] [av]* | 0.88 | 0.89 | 0.89 | 0.67 | 0.65 | 0.49 | 0.86 | 0.67 | 0.67 |
| 0 [non-pol] | 0.93 | 0.9 | 0.91 | 0.35 | 0.99 | 0.52 | 0.72 | 1.00 | 0.83 |
| 1 [pol] | 0.84 | 0.88 | 0.86 | 0.99 | 0.3 | 0.46 | 1.00 | 0.34 | 0.50 |
| *SML [SGD] [av]* | 0.89 | 0.88 | 0.89 | 0.66 | 0.68 | 0.57 | 0.66 | 0.60 | 0.59 |
| 0 [non-pol] | 0.9 | 0.93 | 0.92 | 0.38 | 0.93 | 0.54 | 0.68 | 0.90 | 0.78 |
| 1 [pol] | 0.88 | 0.84 | 0.86 | 0.94 | 0.43 | 0.59 | 0.64 | 0.31 | 0.41 |
| *NN [CNN] [av]* | 0.58 | 0.59 | 0.58 | 0.46 | 0.47 | 0.34 | 0.53 | 0.52 | 0.51 |
| 0 [non-pol] | 0.69 | 0.67 | 0.68 | 0.26 | 0.75 | 0.39 | 0.64 | 0.81 | 0.72 |
| 1 [pol] | 0.48 | 0.50 | 0.49 | 0.65 | 0.19 | 0.29 | 0.43 | 0.23 | 0.30 |
| *NN [LSTM] [av]* | 0.87 | 0.88 | 0.88 | 0.69 | 0.68 | 0.55 | 0.78 | 0.61 | 0.59 |
| 0 [non-pol] | 0.91 | 0.90 | 0.90 | 0.38 | 0.99 | 0.55 | 0.68 | 0.98 | 0.80 |
| 1 [pol] | 0.84 | 0.85 | 0.85 | 0.99 | 0.37 | 0.54 | 0.88 | 0.23 | 0.37 |
| *NN [BERT] [av]* | 0.83 | 0.85 | 0.83 | 0.78 | 0.80 | 0.79 | 0.73 | 0.67 | 0.67 |
| 0 [non-pol] | 0.97 | 0.75 | 0.84 | 0.66 | 0.76 | 0.71 | 0.73 | 0.91 | 0.81 |
| 1 [pol] | 0.70 | 0.96 | 0.81 | 0.90 | 0.84 | 0.87 | 0.74 | 0.43 | 0.54 |

**Table A6.4**. Comparison of techniques for detecting politics-related information (lowercasing/ stemming/no stopwords)

| | TVD | | | DVD | | | WVD | | |
|---|---|---|---|---|---|---|---|---|---|
| | Pr | Rec | F1 | Pr | Rec | F1 | Pr | Rec | F1 |
| *Di-CAP [av]* | 0.72 | 0.73 | 0.70 | 0.66 | 0.70 | 0.63 | 0.76 | 0.77 | 0.77 |
| 0 [non-pol] | 0.87 | 0.62 | 0.72 | 0.43 | 0.84 | 0.57 | 0.85 | 0.78 | 0.81 |
| 1 [pol] | 0.58 | 0.85 | 0.69 | 0.90 | 0.56 | 0.69 | 0.68 | 0.77 | 0.72 |
| *Di-LL [av]* | 0.75 | 0.77 | 0.74 | 0.64 | 0.67 | 0.58 | 0.66 | 0.67 | 0.66 |
| 0 [non-pol] | 0.89 | 0.67 | 0.76 | 0.39 | 0.85 | 0.54 | 0.78 | 0.66 | 0.71 |
| 1 [pol] | 0.62 | 0.86 | 0.72 | 0.89 | 0.48 | 0.63 | 0.54 | 0.68 | 0.61 |
| *Di-CAP-LL [av]* | 0.77 | 0.78 | 0.75 | 0.68 | 0.72 | 0.66 | 0.75 | 0.74 | 0.75 |
| 0 [non-pol] | 0.90 | 0.68 | 0.78 | 0.46 | 0.84 | 0.59 | 0.80 | 0.84 | 0.82 |
| 1 [pol] | 0.63 | 0.88 | 0.73 | 0.91 | 0.61 | 0.73 | 0.71 | 0.64 | 0.67 |
| *SML [PA] [av]* | 0.91 | 0.9 | 0.91 | 0.68 | 0.66 | 0.51 | 0.80 | 0.63 | 0.62 |
| 0 [non-pol] | 0.92 | 0.95 | 0.93 | 0.36 | 0.99 | 0.53 | 0.69 | 0.98 | 0.81 |
| 1 [pol] | 0.91 | 0.86 | 0.88 | 0.99 | 0.33 | 0.5 | 0.90 | 0.28 | 0.42 |
| *SML [BNB] [av]* | 0.88 | 0.89 | 0.88 | 0.67 | 0.64 | 0.48 | 0.83 | 0.57 | 0.51 |
| 0 [non-pol] | 0.92 | 0.91 | 0.91 | 0.35 | 1 | 0.52 | 0.66 | 1.00 | 0.79 |
| 1 [pol] | 0.85 | 0.87 | 0.86 | 1 | 0.28 | 0.44 | 1.00 | 0.13 | 0.23 |
| *SML [MNB] [av]* | 0.9 | 0.91 | 0.9 | 0.7 | 0.74 | 0.64 | 0.82 | 0.65 | 0.64 |
| 0 [non-pol] | 0.95 | 0.89 | 0.92 | 0.44 | 0.96 | 0.6 | 0.70 | 0.99 | 0.82 |
| 1 [pol] | 0.84 | 0.93 | 0.88 | 0.97 | 0.52 | 0.68 | 0.94 | 0.31 | 0.46 |
| *SML [LR] [av]* | 0.89 | 0.9 | 0.89 | 0.67 | 0.64 | 0.48 | 0.84 | 0.66 | 0.66 |
| 0 [non-pol] | 0.92 | 0.92 | 0.92 | 0.35 | 1 | 0.52 | 0.71 | 0.99 | 0.83 |
| 1 [pol] | 0.87 | 0.87 | 0.87 | 1 | 0.28 | 0.43 | 0.97 | 0.33 | 0.49 |
| *SML [SGD] [av]* | 0.88 | 0.88 | 0.88 | 0.66 | 0.68 | 0.58 | 0.64 | 0.58 | 0.57 |
| 0 [non-pol] | 0.91 | 0.91 | 0.91 | 0.39 | 0.9 | 0.55 | 0.67 | 0.89 | 0.77 |
| 1 [pol] | 0.85 | 0.85 | 0.85 | 0.93 | 0.46 | 0.62 | 0.60 | 0.28 | 0.38 |
| *NN [CNN] [av]* | 0.88 | 0.86 | 0.87 | 0.67 | 0.61 | 0.43 | 0.82 | 0.61 | 0.59 |
| 0 [non-pol] | 0.87 | 0.94 | 0.90 | 0.33 | 1.00 | 0.50 | 0.68 | 0.99 | 0.81 |
| 1 [pol] | 0.88 | 0.78 | 0.83 | 1.00 | 0.22 | 0.36 | 0.96 | 0.23 | 0.38 |
| *NN [LSTM] [av]* | 0.82 | 0.83 | 0.83 | 0.65 | 0.66 | 0.54 | 0.63 | 0.61 | 0.61 |
| 0 [non-pol] | 0.88 | 0.85 | 0.87 | 0.37 | 0.93 | 0.53 | 0.69 | 0.83 | 0.76 |
| 1 [pol] | 0.77 | 0.82 | 0.79 | 0.94 | 0.38 | 0.54 | 0.58 | 0.39 | 0.46 |
| *NN [BERT] [av]* | 0.82 | 0.83 | 0.79 | 0.80 | 0.82 | 0.81 | 0.72 | 0.71 | 0.71 |
| 0 [non-pol] | 0.99 | 0.67 | 0.80 | 0.70 | 0.76 | 0.73 | 0.77 | 0.84 | 0.80 |
| 1 [pol] | 0.65 | 0.99 | 0.79 | 0.90 | 0.87 | 0.89 | 0.68 | 0.58 | 0.63 |

**Table A6.5**. Comparison of techniques for detecting politics-related information (lowercasing/lemmatisation)

| | TVD | | | DVD | | | WVD | | |
|---|---|---|---|---|---|---|---|---|---|
| | Pr | Rec | F1 | Pr | Rec | F1 | Pr | Rec | F1 |
| *Di-CAP [av]* | 0.79 | 0.80 | 0.78 | 0.73 | 0.78 | 0.70 | 0.86 | 0.85 | 0.85 |
| 0 [non-pol] | 0.91 | 0.73 | 0.81 | 0.49 | 0.95 | 0.64 | 0.88 | 0.91 | 0.89 |
| 1 [pol] | 0.67 | 0.88 | 0.76 | 0.97 | 0.61 | 0.75 | 0.84 | 0.79 | 0.81 |
| *Di-LL [av]* | 0.70 | 0.71 | 0.69 | 0.59 | 0.60 | 0.52 | 0.70 | 0.70 | 0.70 |
| 0 [non-pol] | 0.83 | 0.64 | 0.72 | 0.34 | 0.78 | 0.48 | 0.78 | 0.77 | 0.77 |
| 1 [pol] | 0.57 | 0.79 | 0.66 | 0.83 | 0.42 | 0.56 | 0.62 | 0.63 | 0.63 |
| *Di-CAP-LL [av]* | 0.78 | 0.79 | 0.77 | 0.69 | 0.74 | 0.66 | 0.84 | 0.83 | 0.83 |
| 0 [non-pol] | 0.91 | 0.70 | 0.79 | 0.46 | 0.87 | 0.60 | 0.86 | 0.90 | 0.88 |
| 1 [pol] | 0.64 | 0.88 | 0.74 | 0.92 | 0.60 | 0.73 | 0.81 | 0.76 | 0.78 |
| *SML [PA] [av]* | 0.92 | 0.91 | 0.91 | 0.67 | 0.66 | 0.51 | 0.78 | 0.64 | 0.63 |
| 0 [non-pol] | 0.92 | 0.95 | 0.94 | 0.36 | 0.99 | 0.53 | 0.70 | 0.97 | 0.81 |
| 1 [pol] | 0.91 | 0.87 | 0.89 | 0.99 | 0.33 | 0.5 | 0.86 | 0.31 | 0.45 |
| *SML [BNB] [av]* | 0.9 | 0.9 | 0.9 | 0.67 | 0.62 | 0.44 | 0.83 | 0.56 | 0.51 |
| 0 [non-pol] | 0.93 | 0.92 | 0.92 | 0.33 | 1 | 0.5 | 0.66 | 1.00 | 0.79 |
| 1 [pol] | 0.87 | 0.89 | 0.88 | 1 | 0.23 | 0.38 | 1.00 | 0.12 | 0.22 |
| *SML [MNB] [av]* | 0.9 | 0.91 | 0.91 | 0.7 | 0.74 | 0.63 | 0.82 | 0.62 | 0.60 |
| 0 [non-pol] | 0.95 | 0.9 | 0.93 | 0.43 | 0.96 | 0.6 | 0.69 | 0.99 | 0.81 |
| 1 [pol] | 0.85 | 0.92 | 0.89 | 0.97 | 0.51 | 0.67 | 0.96 | 0.24 | 0.39 |
| *SML [LR] [av]* | 0.87 | 0.88 | 0.88 | 0.67 | 0.64 | 0.48 | 0.86 | 0.67 | 0.68 |
| 0 [non-pol] | 0.92 | 0.88 | 0.9 | 0.35 | 0.99 | 0.52 | 0.72 | 1.00 | 0.84 |
| 1 [pol] | 0.82 | 0.88 | 0.85 | 0.99 | 0.28 | 0.44 | 1.00 | 0.35 | 0.52 |
| *SML [SGD] [av]* | 0.89 | 0.89 | 0.89 | 0.65 | 0.67 | 0.56 | 0.75 | 0.66 | 0.66 |
| 0 [non-pol] | 0.9 | 0.93 | 0.92 | 0.38 | 0.92 | 0.54 | 0.72 | 0.94 | 0.81 |
| 1 [pol] | 0.88 | 0.84 | 0.86 | 0.93 | 0.42 | 0.58 | 0.79 | 0.38 | 0.51 |
| *NN [CNN] [av]* | 0.88 | 0.85 | 0.86 | 0.66 | 0.60 | 0.41 | 0.78 | 0.60 | 0.57 |
| 0 [non-pol] | 0.86 | 0.95 | 0.90 | 0.33 | 1.00 | 0.49 | 0.68 | 0.98 | 0.80 |
| 1 [pol] | 0.90 | 0.75 | 0.82 | 1.00 | 0.19 | 0.32 | 0.88 | 0.21 | 0.34 |
| *NN [LSTM] [av]* | 0.88 | 0.89 | 0.88 | 0.67 | 0.66 | 0.53 | 0.75 | 0.60 | 0.58 |
| 0 [non-pol] | 0.92 | 0.89 | 0.91 | 0.37 | 0.97 | 0.54 | 0.68 | 0.97 | 0.80 |
| 1 [pol] | 0.84 | 0.88 | 0.86 | 0.97 | 0.36 | 0.52 | 0.82 | 0.23 | 0.37 |
| *NN [BERT] [av]* | 0.86 | 0.88 | 0.87 | 0.78 | 0.82 | 0.79 | 0.78 | 0.64 | 0.63 |
| 0 [non-pol] | 0.96 | 0.82 | 0.88 | 0.63 | 0.84 | 0.72 | 0.70 | 0.97 | 0.81 |
| 1 [pol] | 0.76 | 0.95 | 0.85 | 0.93 | 0.80 | 0.86 | 0.86 | 0.31 | 0.45 |

**Table A6.6**. Comparison of techniques for detecting politics-related information (lowercasing/ lemmatisation/no stopwords)

| | TVD | | | DVD | | | WVD | | |
|---|---|---|---|---|---|---|---|---|---|
| | Pr | Rec | F1 | Pr | Rec | F1 | Pr | Rec | F1 |
| *Di-CAP [av]* | 0.78 | 0.80 | 0.77 | 0.73 | 0.78 | 0.70 | 0.85 | 0.85 | 0.85 |
| 0 [non-pol] | 0.91 | 0.70 | 0.79 | 0.50 | 0.93 | 0.65 | 0.89 | 0.88 | 0.89 |
| 1 [pol] | 0.65 | 0.89 | 0.75 | 0.96 | 0.63 | 0.76 | 0.81 | 0.82 | 0.81 |
| *Di-LL [av]* | 0.70 | 0.71 | 0.69 | 0.58 | 0.59 | 0.51 | 0.70 | 0.70 | 0.70 |
| 0 [non-pol] | 0.84 | 0.63 | 0.72 | 0.34 | 0.76 | 0.47 | 0.78 | 0.78 | 0.78 |
| 1 [pol] | 0.57 | 0.80 | 0.67 | 0.82 | 0.42 | 0.56 | 0.63 | 0.62 | 0.63 |
| *Di-CAP-LL [av]* | 0.77 | 0.78 | 0.75 | 0.58 | 0.59 | 0.51 | 0.82 | 0.82 | 0.82 |
| 0 [non-pol] | 0.92 | 0.66 | 0.77 | 0.34 | 0.76 | 0.47 | 0.87 | 0.87 | 0.87 |
| 1 [pol] | 0.62 | 0.91 | 0.74 | 0.82 | 0.42 | 0.56 | 0.78 | 0.78 | 0.78 |
| *SML [PA] [av]* | 0.92 | 0.91 | 0.91 | 0.67 | 0.66 | 0.51 | 0.78 | 0.63 | 0.61 |
| 0 [non-pol] | 0.92 | 0.95 | 0.93 | 0.36 | 0.99 | 0.53 | 0.69 | 0.98 | 0.81 |
| 1 [pol] | 0.92 | 0.86 | 0.89 | 0.99 | 0.32 | 0.49 | 0.87 | 0.28 | 0.42 |
| *SML [BNB] [av]* | 0.9 | 0.9 | 0.9 | 0.67 | 0.64 | 0.47 | 0.83 | 0.56 | 0.51 |
| 0 [non-pol] | 0.93 | 0.92 | 0.92 | 0.35 | 1 | 0.51 | 0.66 | 1.00 | 0.79 |
| 1 [pol] | 0.87 | 0.89 | 0.88 | 1 | 0.27 | 0.43 | 1.00 | 0.12 | 0.22 |
| *SML [MNB] [av]* | 0.9 | 0.91 | 0.9 | 0.72 | 0.77 | 0.67 | 0.80 | 0.63 | 0.62 |
| 0 [non-pol] | 0.95 | 0.9 | 0.92 | 0.46 | 0.98 | 0.63 | 0.70 | 0.98 | 0.82 |
| 1 [pol] | 0.85 | 0.92 | 0.88 | 0.99 | 0.56 | 0.71 | 0.90 | 0.29 | 0.43 |
| *SML [LR] [av]* | 0.89 | 0.89 | 0.89 | 0.67 | 0.63 | 0.47 | 0.85 | 0.66 | 0.66 |
| 0 [non-pol] | 0.92 | 0.91 | 0.92 | 0.35 | 1 | 0.51 | 0.71 | 1.00 | 0.83 |
| 1 [pol] | 0.86 | 0.88 | 0.87 | 1 | 0.27 | 0.42 | 1.00 | 0.32 | 0.48 |
| *SML [SGD] [av]* | 0.87 | 0.89 | 0.88 | 0.67 | 0.7 | 0.61 | 0.71 | 0.65 | 0.66 |
| 0 [non-pol] | 0.93 | 0.88 | 0.9 | 0.41 | 0.88 | 0.56 | 0.72 | 0.90 | 0.80 |
| 1 [pol] | 0.82 | 0.89 | 0.85 | 0.92 | 0.52 | 0.66 | 0.70 | 0.41 | 0.52 |
| *NN [CNN] [av]* | 0.88 | 0.86 | 0.86 | 0.67 | 0.61 | 0.42 | 0.81 | 0.64 | 0.64 |
| 0 [non-pol] | 0.87 | 0.94 | 0.90 | 0.33 | 1.00 | 0.50 | 0.70 | 0.98 | 0.82 |
| 1 [pol] | 0.88 | 0.77 | 0.83 | 1.00 | 0.21 | 0.35 | 0.91 | 0.31 | 0.46 |
| *NN [LSTM] [av]* | 0.89 | 0.87 | 0.88 | 0.67 | 0.67 | 0.53 | 0.68 | 0.55 | 0.51 |
| 0 [non-pol] | 0.88 | 0.95 | 0.92 | 0.37 | 0.97 | 0.54 | 0.65 | 0.96 | 0.78 |
| 1 [pol] | 0.91 | 0.79 | 0.85 | 0.97 | 0.36 | 0.53 | 0.70 | 0.14 | 0.24 |
| *NN [BERT] [av]* | 0.85 | 0.87 | 0.84 | 0.83 | 0.85 | 0.84 | 0.76 | 0.72 | 0.72 |
| 0 [non-pol] | 0.98 | 0.76 | 0.86 | 0.74 | 0.81 | 0.77 | 0.76 | 0.89 | 0.82 |
| 1 [pol] | 0.72 | 0.98 | 0.83 | 0.92 | 0.89 | 0.90 | 0.75 | 0.54 | 0.63 |